\documentclass{article}

\usepackage[preprint]{neurips_2023}

\usepackage[utf8]{inputenc} 
\usepackage[T1]{fontenc}    
\usepackage{hyperref}       
\usepackage{url}            
\usepackage{booktabs}       
\usepackage{amsfonts}       
\usepackage{nicefrac}       
\usepackage{microtype}      
\usepackage{xcolor}         
\usepackage{float}
\usepackage{graphicx}
\usepackage{amsmath}
\usepackage{subcaption}

\title{A Foundation Model for Soccer}
\author{Ethan Baron\thanks{Equal contribution.} \\ University of Toronto \And Daniel Hocevar$^*$ \\ University of Toronto \And Zach Salehe$^*$ \\ University of Toronto}

\begin{document}

\maketitle

\begin{abstract}
We propose a foundation model for soccer, which is able to predict subsequent actions in a soccer match from a given input sequence of actions. As a proof of concept, we train a transformer architecture on three seasons of data from a professional soccer league. We quantitatively and qualitatively compare the performance of this transformer architecture to two baseline models: a Markov model and a multi-layer perceptron. Additionally, we discuss potential applications of our model. We provide an open-source implementation of our methods at \url{https://github.com/danielhocevar/Foundation-Model-for-Soccer}.
\end{abstract}

\section{Introduction}

In recent years, adapting deep learning models trained using self-supervised learning on large datasets has proven effective on a variety of downstream tasks. These large pre-trained models, called ``foundation models'', have had a profound impact on fields such as natural language processing and computer vision \citep{foundation_models}.

We propose to train a foundation model for play-by-play data in soccer based on action data from historical matches. Much like language foundation models are trained to predict the next word given an input context, our soccer foundation model will be trained to predict the subsequent action given an input sequence of actions in a soccer match. Thereby, our model will learn embedding representations for soccer actions that could be used in a variety of downstream tasks. Specifically, our foundation model's embedding representation for actions and sequences of actions could be used as inputs to other models, or as a seed from which to simulate subsequent actions.

There are a variety of interesting use cases for our model. Firstly, the foundation model could be used to generate sequences of future actions given an input sequence, allowing analysts to simulate possible progressions of a match from a given situation. This could allow analysts to analyze tactical decisions within matches. Secondly, the hidden representations of the foundation model can themselves be used as inputs to other models, enabling analysts to refine data-driven player evaluation pipelines.

\section{Background}

\subsection{Deep Learning for Sequence Modeling}


One of the largest breakthroughs in recent years with respect to sequence modeling was the introduction of the transformer model \citep{transformer}. A transformer is a neural architecture that transforms an input sequence into an output sequence. The driving idea behind transformers is the attention mechanism. At each step in the processing, this mechanism decides which other parts of the input sequence are most important for predicting the subsequent word. Transformers are widely applicable in a variety of downstream tasks. For example, this model could be trained to translate text (sequences of words) from one language to another, or for creating a chat bot.

Transformers come in several different flavours as well. An encoder-only transformer architecture only features encoding layers and omits any sort of decoding process. Encoder-only models are widely useful in natural language understanding and in creating highly valuable text embeddings which can later be applied to a variety of downstream tasks, such as text classification and search. A notable encoder-only model is Google's BERT \citep{bert}, which features a bidirectional transformer, allowing it to make use of both preceding and succeeding words when analyzing a sequence. Conversely, decoder-only transformer architectures only feature decoding layers, and their main purpose lies in generating new words or sequences based on the inputted representations. Popular decoder-only models include OpenAI's GPT models \citep{gpt2}. Encoder-decoder models, as seen in the original transformer paper \citep{transformer}, feature both encoders and decoders. These models are commonly used for sequence-to-sequence mapping tasks, such as translating text from one language to another.


\subsection{Modeling Soccer Actions}

\citet{seq2event} were the first to apply natural language processing architectures such as RNNs and transformers to predict soccer actions. They train multiple different architectures to predict the coordinates and action type of the subsequent action, given an input sequence of actions. They frame this problem as a multi-objective problem, with a regression component to predict the coordinates of the subsequent action, and a classification component to predict the action type. They found that both transformers and LSTMs offered significant improvements in predictive power over a Markov baseline model. However, they do not consider sequences that include turnovers, where the possession of the ball changes from one team to the other. In contrast to their work, we tokenize actions allowing us to treat the task as a pure classification problem, and we also consider all sequences from a soccer match, including turnovers.

\cite{soccer_lem} construct a foundation model for soccer actions which they call a Large Action Model. They use a public dataset of actions from soccer matches provided by Wyscout and describe various downstream tasks relying on this foundation model, such as for an in-game win probability model. Their architecture is a multi-layer perceptron trained to predict the next action given an input sequence of the three previous actions. In contrast to their work, we use a deep learning architecture that is more well-suited to the sequential nature of soccer actions, and we also consider longer input sequences.

\citet{football2vec} apply the ideas from \texttt{word2vec} \citep{word2vec} to learn embeddings for soccer actions by leveraging co-occurrences of actions. They tokenize actions into a standardized textual format, specifying the action type, location (in a 5x5 grid), and any other additional information. For example, the token ``(1/5,1/5)$<$dribble$>$: incomplete" refers to ``an unsuccessful dribble on the left side of the defense". The action embeddings are then used to identify similar actions and develop notions of player style. In contrast to this work, we use our trained embeddings as inputs to a generative model and learn representations for sequences of multiple actions.

\section{Methods}
\subsection{Dataset}

We use a tabular play-by-play dataset of matches from the 2018-2019, 2019-2020, and 2020-2021 seasons of the FA Women's Super League, widely considered one of the top women's soccer leagues in the world. The data is provided as part of the \href{https://github.com/statsbomb/open-data}{StatsBomb Open Data repository}, a collection of action data for various historical soccer competitions made publicly available by StatsBomb, an industry-leading data provider. Our dataset includes a total of 939920 actions from 326 matches across the three seasons. To ensure that the representations learned by our model can generalize across different teams and seasons, we evaluate our model's performance on a withheld test set containing 10\% of the matches. We also use 10\% of the matches as a dedicated validation set, which is used for hyperparameter tuning.

For data collection and pre-processing, we leverage the \href{https://github.com/ML-KULeuven/socceraction}{\texttt{socceraction}} Python package. This package includes functionality to convert the raw data into a representation called SPADL (Soccer Player Action Description Language) \citep{decroos2019actions}, a standardized representation which focuses on player-centric actions and a unified set of features. We focus our analysis on the following features:
\begin{enumerate}
    \item \texttt{team}: which team the action is associated with (home or away)
    \item \texttt{action type}: what type of action occurred (e.g. ``pass'', ``dribble'')
    \item \texttt{x} and \texttt{y}: the $(x, y)$ coordinates of the action on the soccer field, in yards
\end{enumerate}

We choose to model actions in a discrete space, which allows us to apply techniques from natural language processing. We must therefore discretize the $(x,y)$ coordinates of the actions. To do so, we split up the soccer field, which has dimensions of 105 $\times$ 68 yards, into a hundred equally-sized rectangles. For example, rectangle $(0, 0)$ corresponds to the range $x \in[0, 10.5), y \in [0, 6.8)$ and rectangle $(1, 2)$ corresponds to the range $x \in [10.5, 21), y \in [13.6, 20.4)$. We can then tokenize each action into the following format: ``<team>, <action type>, <bin>''. For instance, the token ``True, dribble, 4, 4'' indicates a dribble by the home team in bin $(4, 4)$, which is near the middle of the field.

\subsection{Neural Network Architecture}

We frame the task of training a soccer action sequence embedding model as an unsupervised learning task, as in \citet{unsupervised}. Specifically, given a corpus of actions $S = a_1, a_2, \ldots, a_n$, the model is tasked with predicting the subsequent action $a_t$. Using the discretized action space described before, we minimize the cross-entropy loss using the log softmax of the transformer's output logits. This objective is equivalent to maximizing the likelihood of the data as follows:
\begin{equation}\label{eq:loss}
    L(S; w) = \sum_{i=k+1}^n \log P(a_i | a_{i-k}, \ldots, a_{i-1}; w)
\end{equation}

\begin{figure}[h]
    \begin{center}
    \includegraphics[width=\textwidth]{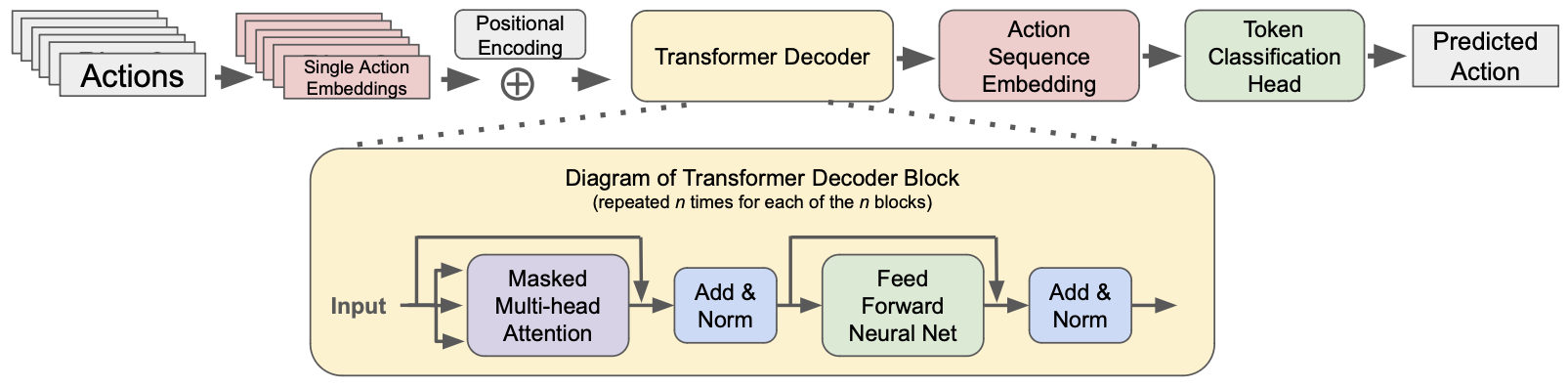}
    \end{center}
    \caption{Model architecture diagram}
    \label{architecture_diagram}
\end{figure}

Figure \ref{architecture_diagram} illustrates the architecture of our transformer model. Each individual action is first converted to an embedding representation. The sequence of action embeddings are then combined with a positional encoding, and passed into a transformer decoder, following the architecture described by \cite{unsupervised}. This transformer decoder consists of $n$ blocks which learn an embedding representing the sequence of actions. Finally, in order to produce predictions for the learning task described in Equation \ref{eq:loss}, we pass the transformer-learned representation for the last input token into a token classification head, which is implemented as a single fully-connected layer. This token classification head generates logits for each of the possible tokens, which are then passed through a softmax function to arrive at the predicted probabilities.

\emph{\textbf{Hyperparameters}} We use an action embedding dimension of 50. In each block of the transformer decoder, we use two layers for the feed-forward neural network with a hidden layer size of 500 and ReLu activation. For the token classification head, we simply use a fully-connected linear layer. For the positional encoding, we use sinusoidal functions of various frequencies, as proposed by \citet{transformer}.

\emph{\textbf{Optimization}} We use an Adam optimizer with a learning rate of 0.0002 and batch size of 32. We train the model for 6 epochs, and omit the use of dropout regularization within our positional encoding module.

\emph{\textbf{Configuration}} We introduce two different sizes of the model described in Figure \ref{architecture_diagram}, each with a different number of decoder layers and attention heads. The specifications for these models are outlined in the table below:

\begin{table}[h]
\centering\caption{Comparison of the two model sizes}
\label{models}
\begin{tabular}{l|c|c|c}
\hline
Model & Attention Heads & Decoder Blocks & Total Parameters \\
\hline
Small & 2 & 1 & 365,565\\
Large & 5 & 4 & 548,415\\
\hline
\end{tabular}
\end{table}

\subsection{Baseline Models}

\emph{\textbf{Markov Model}} For our first baseline model, we implement a 2-gram language model, which essentially uses the observed pairs of actions in the training data to estimate the probability of each subsequent action from the given previous action. This approach encodes a Markov assumption underlying the data-generating process, which is probably too simple for our domain. However, it is easy to implement, and has reasonable inductive biases, hence it is a good baseline to compare against.

We train the model using observed sequences of actions in the training data $a_1, a_2, \ldots, a_n$. We construct a transition count matrix $T$, where element $(i,j)$ indicates the number of times in the training data that token $j$ was followed by token $i$. To avoid probabilities of zero, we introduce Laplace smoothing by adding 0.1 to each element. That is,
\[ T_{ij} = 0.1 + \sum_{t=1}^{n-1} \mathbb{I}[a_t = i] \mathbb{I}[a_{t+1} = j]\]
After constructing this transition count matrix, we normalize each row to sum to one, giving us the transition probability matrix $P$:
\[ P_{ij} = \frac{T_{ij}}{\sum_k T_{ik}}\]
Intuitively, this baseline Markov model simply learns to reproduce the observed frequencies of pairs of actions in the training set.

\emph{\textbf{Multilayer Perceptron}} For our second baseline model, we replace the transformer decoder and token classification head with a simple MLP network, similar to the approach in \citet{soccer_lem}. That is, each individual action first gets converted into an embedding representation, and the sequence is then positionally encoded. Following this, we concatenate the sequence into a single vector, which is passed through the MLP to predict the following action probabilities.

We use an embedding dimension of 128, along with omitting the use of dropout. Our MLP consists of 4 fully-connected layers, with each layer having 1024 hidden features. We use ReLU activation and apply a softmax activation to the final output. When training, we use an Adam optimizer with a learning rate of 0.0002 and a batch size of 100, which we run for 3 epochs.

\section{Results}

\subsection{Quantitative Results}
\begin{table}[h]
\centering\caption{Comparison of quantitative results (TSFR indicates transformer)}
\label{results}
\begin{tabular}{l|c|c|c|c|c|c|c|c}
\hline
 & \multicolumn{4}{c|}{Accuracy} & \multicolumn{4}{c}{Mean Log Likelihood} \\ \hline
Dataset & Markov & MLP & TF (S) & TF (L) & Markov & MLP & TF (S) & TF (L) \\
\hline
Train & 0.420 & 0.442 & 0.425 & 0.429 & -2.847 & -2.342 & -2.469 & -2.447 \\
Val & 0.417 & 0.411 & 0.421 & 0.422 & -3.016 & -2.674 & -2.587 & -2.574 \\
Test & 0.424 & 0.418 & 0.428 & 0.429 & -2.975 & -2.632 & -2.549 & -2.534 \\
\hline
\end{tabular}
\end{table}

To evaluate the performance of our models quantitatively, we consider two metrics. First, we compute the models' accuracy on each subset of the data, measuring the models' ability to predict the most likely next action from a given situation. Second, we compute the mean log likelihood per example on each subset of the data, measuring how well-calibrated the models' predicted probabilities are. This metric is equivalent to the negative of the mean cross entropy loss per example.

We present the quantitative results of our models in Table \ref{results}. We see that both sizes of the transformer architecture achieved slightly higher accuracy than the Markov model for all three datasets. This suggests that the Markov model is adequately identifying the most likely next action given an input sequence. However, we see that the transformer architecture does achieve a significantly higher mean log likelihood than the Markov model, suggesting that the next action probabilities predicted by the transformer model are more well-calibrated. While the MLP also achieves a higher mean log likelihood than the Markov model, it falls short of both the large and small variants of the transformer when it comes to mean log likelihood over the validation and test sets. For this reason, we choose to utilize a variant of the transformer model for the rest of our analysis. In this case, the large variant of the transformer model outperforms the small transformer variant across both validation accuracy and mean log likelihood metrics, indicating its predictions are the most accurate and well-calibrated of all the models we tested. 

\subsection{Transformer Scaling Laws}
\begin{figure}[h]
    \begin{center}
    \includegraphics[width=\textwidth]{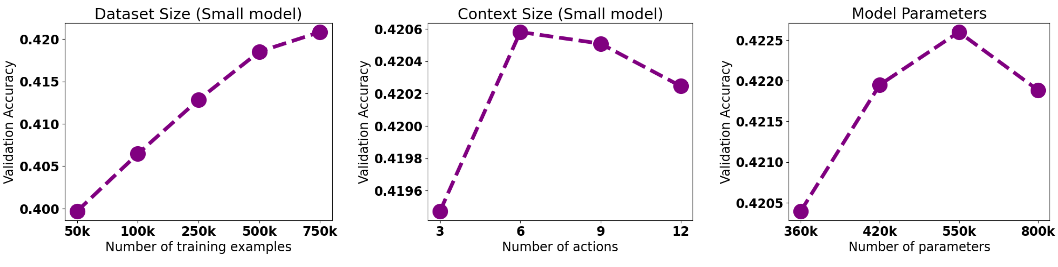}
    \end{center}
    \caption{Plots showing how the validation accuracy of the model varies depending on dataset size, context size and number of parameters}
    \label{scaling}
\end{figure}

Following the approach of \cite{scaling}, we measure how our model's validation accuracy scales depending on the size of the training dataset, the number of actions in the model's context window, and the total number of parameters in the model. The results of this analysis are summarized in Figure \ref{scaling}. These results can be interpreted in order to provide insight into how the model can be extended or improved in order to achieve a better predictive accuracy. For computational reasons, we perform this analysis using the ``Small'' version of the transformer architecture.

Given that the validation accuracy of the model improves as the number of training examples increases, gaining access to a larger dataset than the one we use now would likely improve the accuracy of the model. However, we find that there are diminishing returns to increasing the context window of the model, as well as the number of parameters. Specifically, increasing the context size beyond 9 and the number of parameters beyond 550k does not appear to improve the accuracy of the model. These findings likely indicate that a more complex model is more likely to overfit, making it harder to achieve satisfactory performance on unseen data.

\subsection{Visualizing Embeddings}

\begin{figure*}[h]
    \centering
    \begin{subfigure}[b]{0.475\textwidth}
        \centering
        \includegraphics[width=\textwidth]{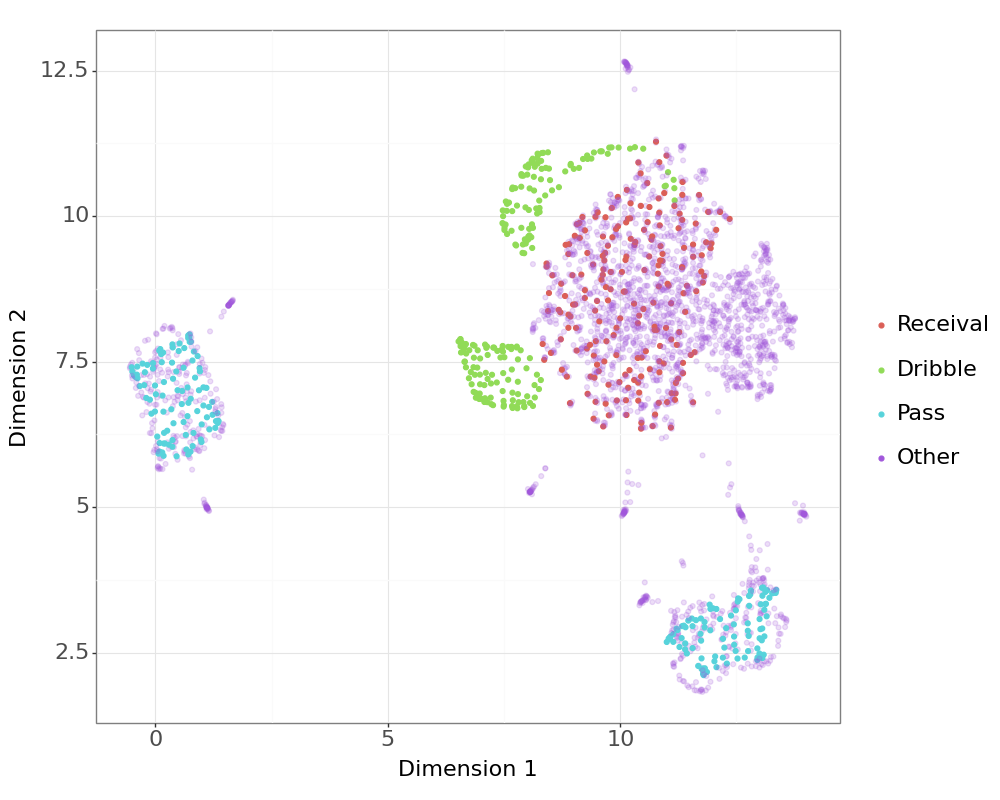}
        \caption[]%
        {{\small Embeddings by action type}}    
    \end{subfigure}
    \hfill
    \begin{subfigure}[b]{0.475\textwidth}  
        \centering 
        \includegraphics[width=\textwidth]{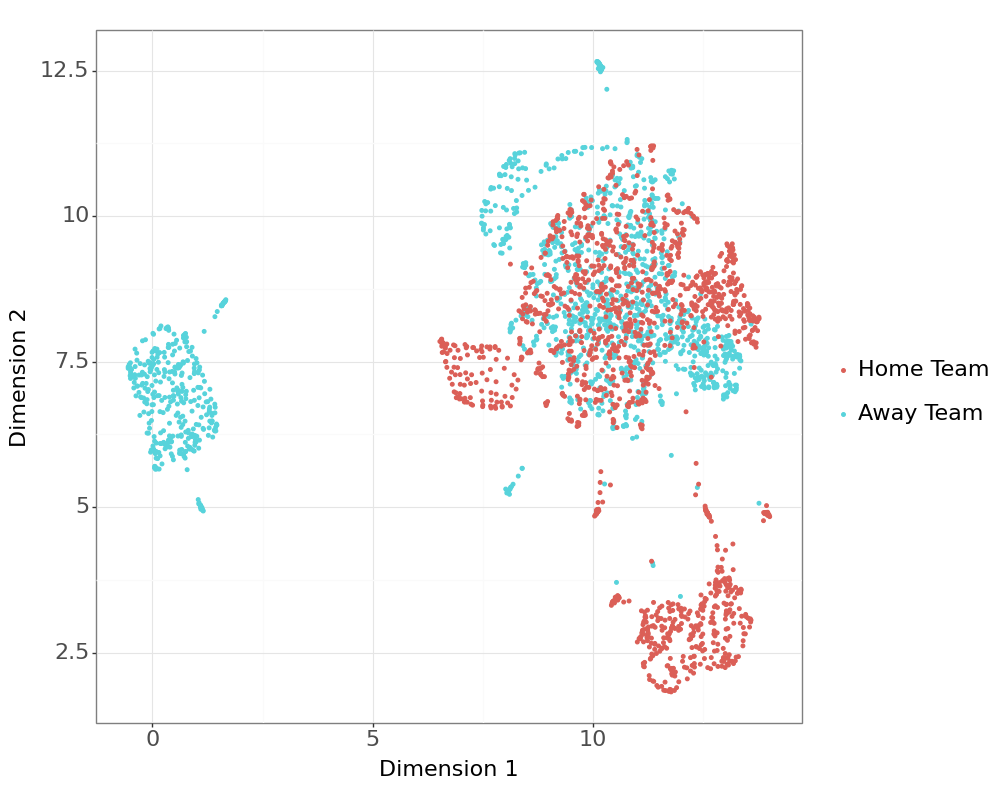}
        \caption[]%
        {{\small Embeddings by team}}    
    \end{subfigure}
    \vskip\baselineskip
    \begin{subfigure}[b]{0.475\textwidth}   
        \centering 
        \includegraphics[width=\textwidth]{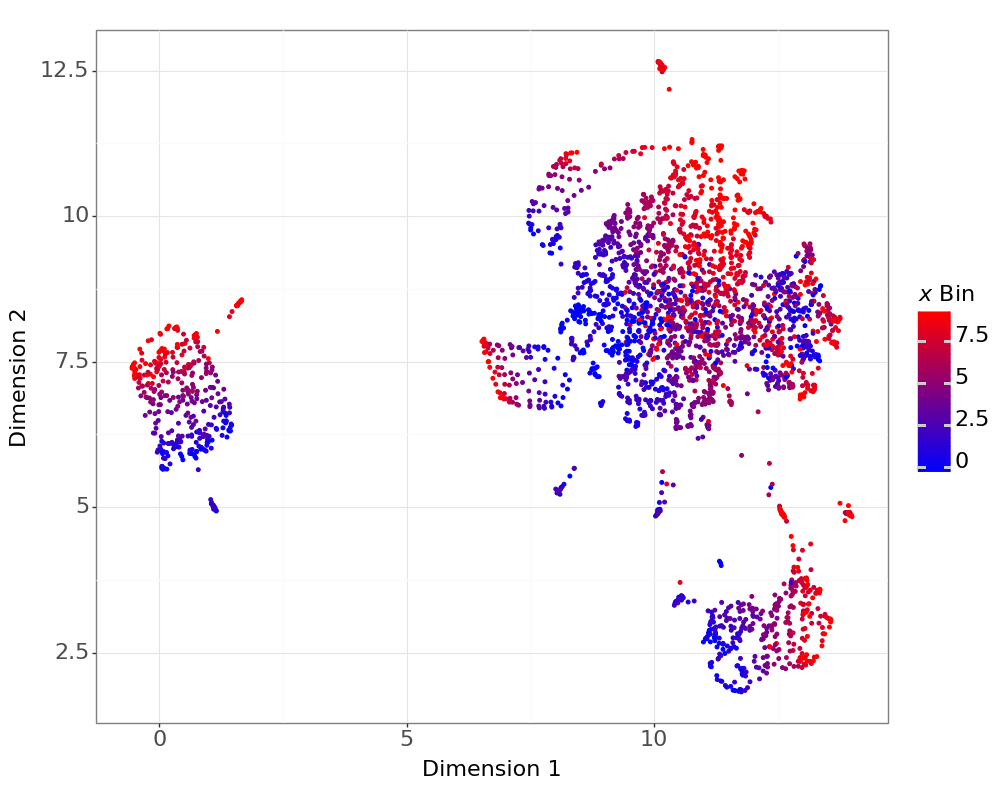}
        \caption[]%
        {{\small Embeddings by x-coordinate}}    
    \end{subfigure}
    \hfill
    \begin{subfigure}[b]{0.475\textwidth}   
        \centering 
        \includegraphics[width=\textwidth]{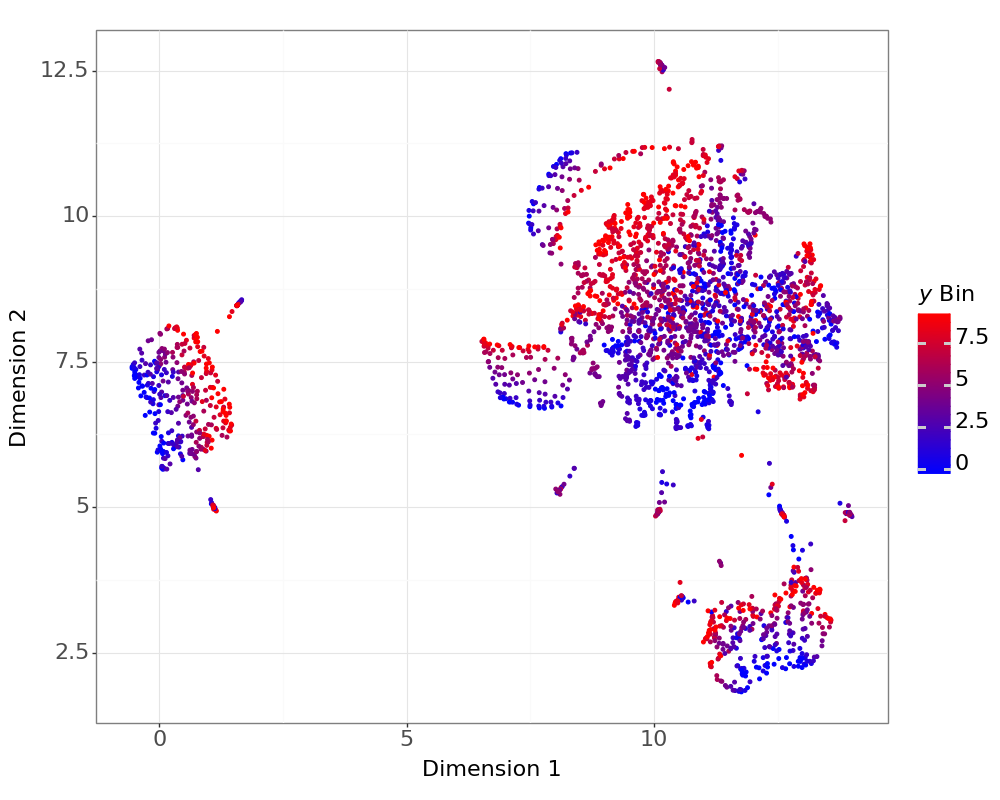}
        \caption[]%
        {{\small Embeddings by y-coordinate}}    
    \end{subfigure}
    \caption[]
    {\small Visualizations of individual play embeddings} 
    \label{emb}
\end{figure*}

Figure \ref{emb} provides a visualization of the learned embeddings for individual plays in out dataset. We use UMAP \citep{umap} to reduce our 50-dimensional action embeddings into a two-dimensional space for visualization. We see that the model has automatically learned to group similar actions together, such as having distinct clusters for certain action types for each team. Furthermore, the model has learned to represent the geometry of the soccer field, as we see consistent patterns with respect to the bin labels.

\subsection{Example Outputs}

\begin{figure*}[h]
    \centering
    \begin{subfigure}[b]{0.475\textwidth}
        \centering
        \includegraphics[width=\textwidth]{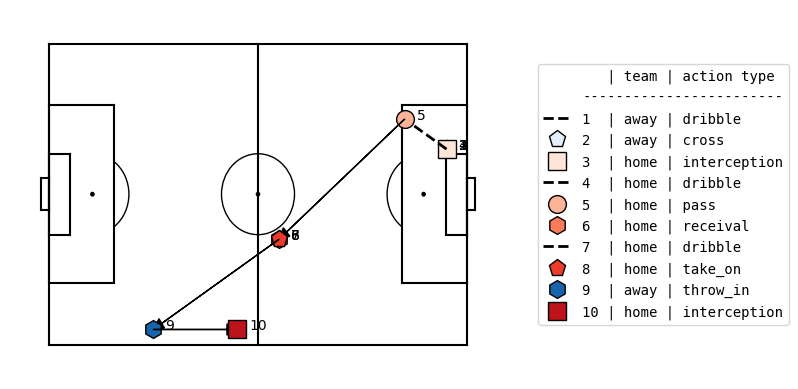}
        \caption[]%
        {{\small Ground truth}}    
    \end{subfigure}
    \hfill
    \begin{subfigure}[b]{0.475\textwidth}  
        \centering 
        \includegraphics[width=\textwidth]{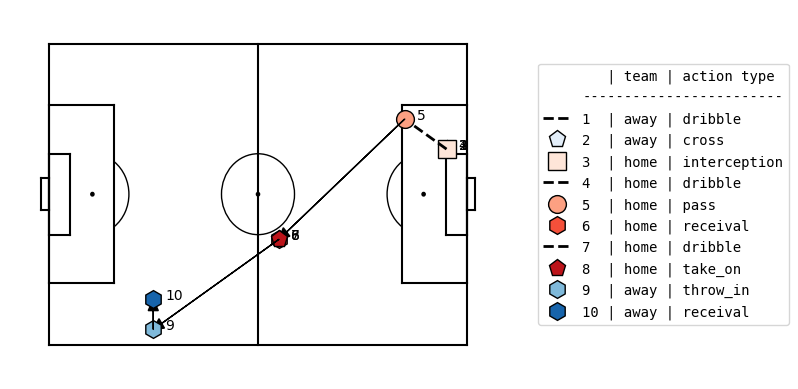}
        \caption[]%
        {{\small Markov}}    
    \end{subfigure}
    \vskip\baselineskip
    \begin{subfigure}[b]{0.475\textwidth}   
        \centering 
        \includegraphics[width=\textwidth]{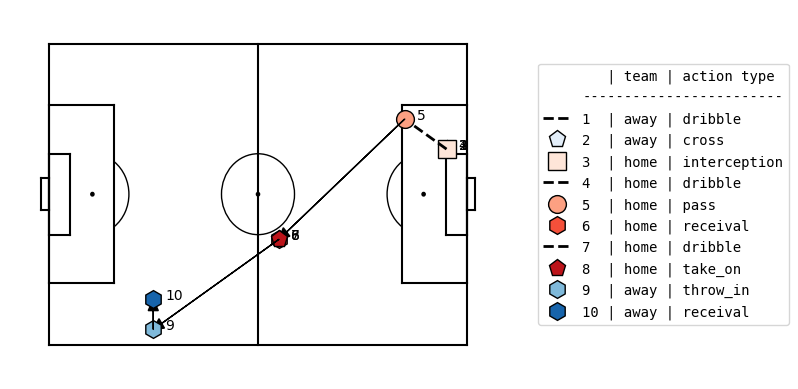}
        \caption[]%
        {{\small MLP}}    
    \end{subfigure}
    \hfill
    \begin{subfigure}[b]{0.475\textwidth}   
        \centering 
        \includegraphics[width=\textwidth]{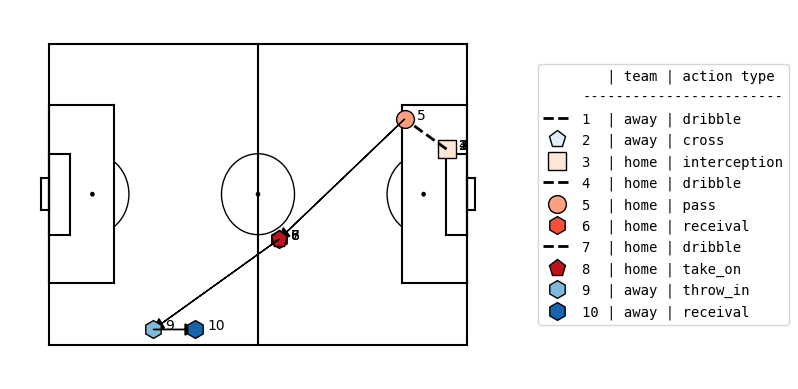}
        \caption[]%
        {{\small Transformer (large)}}    
    \end{subfigure}
    \caption[]
    {\small Sequence where models fail} 
    \label{soccer_plots_bad_1}
\end{figure*}

\begin{figure*}[h]
    \centering
    \begin{subfigure}[b]{0.475\textwidth}
        \centering
        \includegraphics[width=\textwidth]{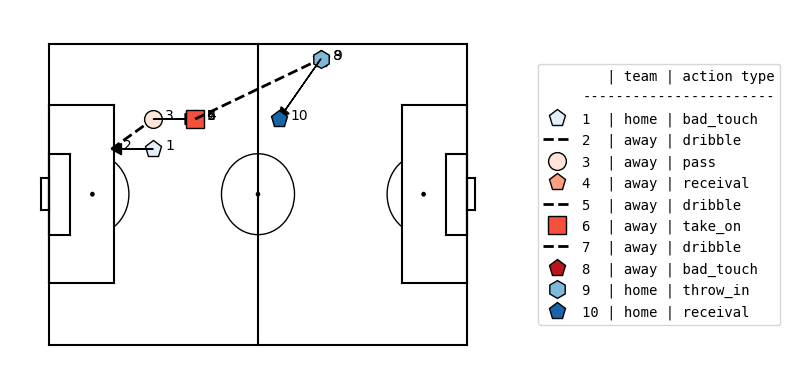}
        \caption[]%
        {{\small Ground truth}}    
    \end{subfigure}
    \hfill
    \begin{subfigure}[b]{0.475\textwidth}  
        \centering 
        \includegraphics[width=\textwidth]{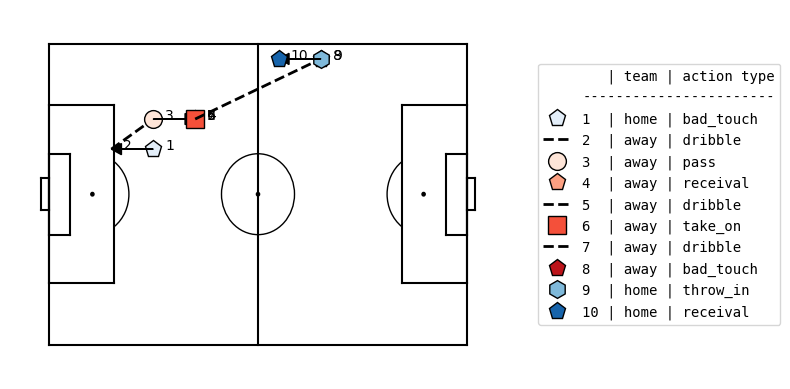}
        \caption[]%
        {{\small Markov}}    
    \end{subfigure}
    \vskip\baselineskip
    \begin{subfigure}[b]{0.475\textwidth}   
        \centering 
        \includegraphics[width=\textwidth]{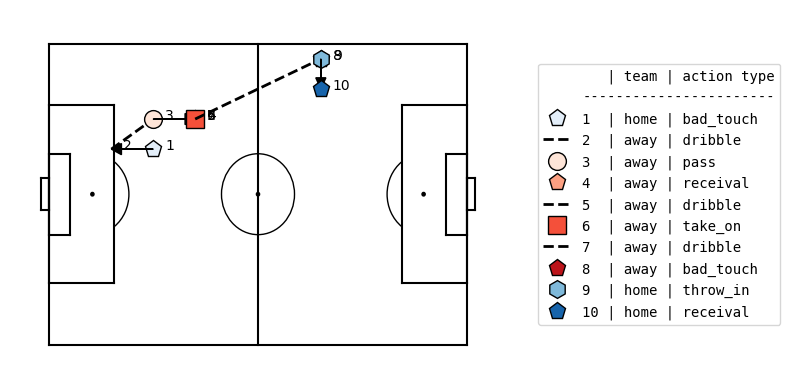}
        \caption[]%
        {{\small MLP}}    
    \end{subfigure}
    \hfill
    \begin{subfigure}[b]{0.475\textwidth}   
        \centering 
        \includegraphics[width=\textwidth]{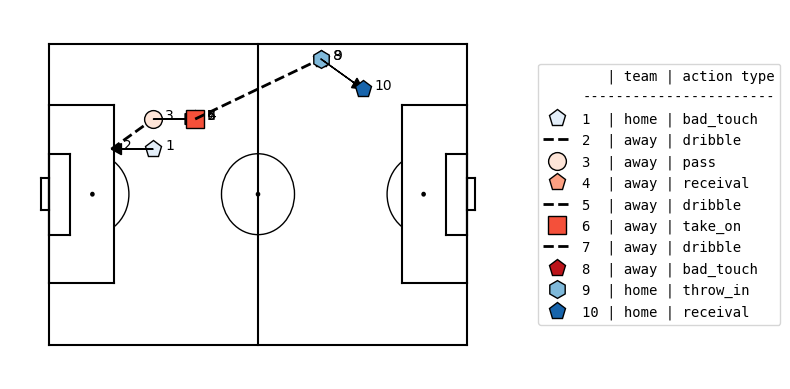}
        \caption[]%
        {{\small Transformer (large)}}    
    \end{subfigure}
    \caption[]
    {\small Another sequence where models fail} 
    \label{soccer_plots_bad_2}
\end{figure*}

\begin{figure*}[h]
    \centering
    \begin{subfigure}[b]{0.475\textwidth}
        \centering
        \includegraphics[width=\textwidth]{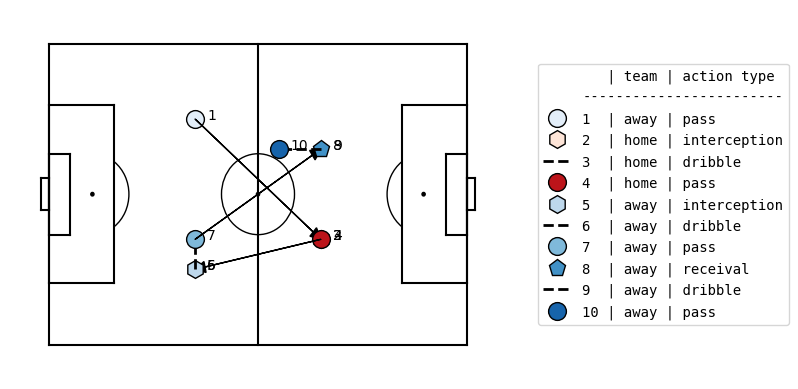}  
    \end{subfigure}
    \hfill
    \begin{subfigure}[b]{0.475\textwidth}  
        \centering 
        \includegraphics[width=\textwidth]{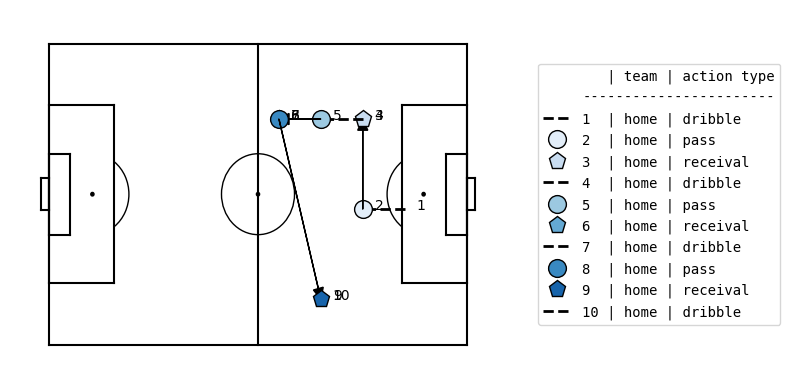}
    \end{subfigure}
    \caption[]
    {\small Two sequences where models succeed} 
    \label{soccer_plots_good}
\end{figure*}

Figures \ref{soccer_plots_bad_1}, \ref{soccer_plots_bad_2}, and \ref{soccer_plots_good} all show visual examples of a 10-play soccer sequence. 

Figures \ref{soccer_plots_bad_1} and \ref{soccer_plots_bad_2} demonstrate the full ground truth sequence, along with the same sequence, but with the last play being predicted by each of the 3 models, given the previous 9 plays as input. These plots in particular demonstrate a couple of failure cases between each of the models. From inspecting a variety of failure cases, it becomes apparent that all 3 models tend to fail on the same sequences way more often than not. In figure \ref{soccer_plots_bad_1} for example, the models fail at correctly predicting the occurrence of an interception play. This makes sense, as interceptions are less frequent than other play types, and can occur out of nowhere, leading them to be hard to predict in general. In figure \ref{soccer_plots_bad_2}, the models are able to recognize that the play following a throw in should be the receival of the throw in. However, none of the models are able to predict where the throw in will be received; understandably so, since in practice this has little to do with the previous actions and more to do with where each player is on the field.

Figure \ref{soccer_plots_good} shows two examples of where all models successfully predict the correct subsequent play. For each sequence in the figure, only one plot is shown, as each model's output was identical to the ground truth.

\section{Potential Applications}

Our foundation model could be used in a variety of interesting applications.

Firstly, the generative nature of our model allows analysts to simulate sequences of soccer actions from a given input sequence. This functionality allows teams to analyze tactical decisions in different situations, and identify the most successful strategies to pursue in a given context. For example, by simulating many sequences of actions following a given input, analysts could determine whether passing the ball to a given area of the pitch is likely to result in more positive outcomes for the team than passing to another area of the pitch.

Secondly, the sequence embedding learned by our model could be used in several downstream applications. These applications include using the representation of a given state within a soccer match to build an in-game match win probability model, or a continuous indicator for which team has the momentum in a match, as in \citet{soccer_lem}. Another interesting idea is to include this multivariate state representation as an input to an expected possession value model \citep{epv}.

Thirdly, the action and sequence embeddings from our model could be used to develop an improved understanding of the unique strength and style of individual soccer players and teams. Just as \citet{football2vec} uses the average embedding of all of a player's actions, we could perform a similar aggregation to arrive at understanding of each player's individual role and preferences. Furthermore, one could even expand our architecture to learn player embeddings as part of the model, automatically extracting useful features that indicate differences in tendencies across different players.




\section{Conclusion}
We use a transformer decoder architecture to develop a foundation model for soccer actions, capable of predicting the next action following from a given input sequence. We show that this transformer architecture provides more accurate and well-calibrated predictions than a baseline Markov model, as well as an MLP model. Our foundation model offers many interesting directions for future research, including various applications within soccer analytics.



\newpage
\bibliography{report}


\end{document}